\documentclass[sigconf]{acmart}
\AtBeginDocument{%
  }

\setcopyright{acmlicensed}
\copyrightyear{2025}
\acmYear{2025}
\acmDOI{XXXXXXX.XXXXXXX}
\acmConference[KDD '25]{ACM SIGKDD Conference on Knowledge Discovery and Data Mining}{August 03--07,
  2025}{Toronto, CA}
\acmISBN{978-1-4503-XXXX-X/2018/06}

\usepackage{hyperref}       % hyperlinks
\usepackage{url}            % simple URL typesetting
\usepackage{multirow}
\usepackage{graphicx}
\usepackage{tcolorbox}
\usepackage{subcaption}

\usepackage{xcolor}

\newcommand{\prompt}[1]{%
  \begin{tcolorbox}[%
    colback=gray!10,    % Background color
    colframe=gray,      % Frame color
    fontupper=\small,     % Font size
    title= Prompt template with placeholder     % Title of the box
  ]
    #1
  \end{tcolorbox}%
}

\newcommand{\eval}[1]{%
  \begin{tcolorbox}[%
    colback=orange!10,    % Background color
    colframe=orange,      % Frame color
    fontupper=\small,     % Font size
    title=NEPA expert evaluation of GPT-4 generated triplet  % Title of the box
  ]
    #1
  \end{tcolorbox}%
}

\newcommand{\context}[1]{%
  \begin{tcolorbox}[%
    colback=pink!10,    % Background color
    colframe=pink,      % Frame color
    fontupper=\small,     % Font size
    title=Selected Excerpt  % Title of the box
  ]
    #1
  \end{tcolorbox}%
}

\newcommand{\goodeval}[1]{%
  \begin{tcolorbox}[%
    colback=teal!10,    % Background color
    colframe=teal,      % Frame color
    fontupper=\small,     % Font size
    title=NEPA expert evaluation of GPT-4 generated triplet  % Title of the box
  ]
    #1
  \end{tcolorbox}%
}

\newcommand{\nepaquad}{NEPAQuAD v1.0~}
\newcommand{\maple}{MAPLE~}

\usepackage{glossaries}
\newacronym{llm}{LLM}{Large Language Model}
\newacronym{rag}{RAG}{Retrieval Augmented Generation}
\newacronym{nlp}{NLP}{Natural Language Processing}
\newacronym{eis}{EIS}{Environmental Impact Statement}
\newacronym{nepa}{NEPA}{National Environment Policy Act}

\begin{document}

% \title{MAPLE \& NEPAQuAD: An Essential Duo to Facilitate Faster Environmental Assessment}
% \title{NEPAQuAD \& MAPLE: Evaluating LLMs' Ability to Accelerate Environmental Permitting}
\title{Benchmarking LLMs for Environmental Review and Permitting}

\author{%
  Rounak Meyur$^1$, Hung Phan$^1$$^,$$^2$, Koby Hayashi$^1$, Ian Stewart$^1$, Shivam Sharma$^1$,
  Sarthak Chaturvedi$^1$, \\ Mike Parker$^1$, Dan Nally$^1$, Sadie Montgomery$^1$, Karl Pazdernik$^1$, 
  Ali Jannesari$^2$, \\ Mahantesh Halappanavar$^1$, Sai Munikoti$^1$, Sameera Horawalavithana$^1$, Anurag Acharya$^1$\\
  \textit{$^1$Pacific Northwest National Laboratory,
  Richland, WA} \\
  \textit{$^2$Iowa State University, Ames, IA}
}

\renewcommand{\shortauthors}{Meyur et al.}

\renewenvironment{quote}
  {\small\list{}{\rightmargin=0.25cm \leftmargin=0.25cm}%
   \item\relax}
  {\endlist}

\begin{abstract}
% \sameera{Please change the title, it is bad.}
% Environmental Impact Statements (EISs) prepared under the \gls*{nepa} are important federal documents that disclose the potential environmental impact of proposed projects.  
% Preparing EIS data involves synthesizing \gls*{nepa} literature spanning several decades, making them time-intensive components of federal project assessments. 
The \gls*{nepa} stands as a foundational piece of environmental legislation in the United States, requiring federal agencies to consider the environmental impacts of their proposed actions. 
The primary mechanism for achieving this is through the preparation of Environmental Assessments (EAs) and, for significant impacts, comprehensive \gls*{eis}.
\gls*{llm}s' effectiveness in specialized domains like \gls*{nepa} remains untested for adoption in federal decision making processes. To address this gap, we present \textbf{NEPA} \textbf{Qu}estion and \textbf{A}nswering \textbf{D}ataset (\textit{NEPAQuAD}), the first comprehensive benchmark derived from EIS documents, along with a modular and transparent evaluation pipeline \textit{\maple} to assess \gls*{llm} performance on NEPA focused regulatory reasoning tasks. 
Our benchmark leverages actual \gls*{eis} documents to create diverse question types, ranging from factual to complex problem-solving ones. 
We built a modular and transparent evaluation pipeline to test both closed- and open-source models in zero-shot or context-driven QA benchmarks.
% Our evaluation pipeline, MAPLE, is built to be modular such that it can be used for any closed- or open-source models on any QA benchmark with any or no context provided, allowing users to evaluate various models on their domain in a single run. 
We evaluate five state-of-the-art \gls*{llm}s -- Claude Sonnet 3.5, Gemini 1.5 Pro, GPT-4, Llama 3.1, and Mistral-7B-Instruct -- using our framework to assess both their prior knowledge and their ability to process \gls*{nepa}-specific information. 
The experimental results reveal that all the models consistently achieve their highest performance when provided with the gold passage as context. While comparing the other context-driven approaches for each model, \gls*{rag}-based approaches substantially outperform PDF document contexts indicating that neither model is well suited for long-context question-answering tasks.
% We observe that all the models, regardless of the context supplied to help answer the questions, perform in the range of 60--75\% correctness, with only one model in one setting achieving correctness of over 80\%.
% \sameera{can we change the main observation to something more appealing than just presenting number improvement, E.g., LLM struggles in NEPA document, long context understanding, etc.}
Our analysis suggests that \gls*{nepa} focused regulatory reasoning tasks pose a significant challenge for \gls*{llm}s, particularly in terms of understanding the complex semantics and effectively processing the lengthy regulatory documents.
% These results show that while the models are able to perform reasonably well, there is still a lot of progress to be made.

% \note{needs to be rewritten to fit the new narrative of `collaborative reasoning'}

\end{abstract}


\received{18 May 2025}

\maketitle

\section{Introduction}
\label{sec:introduction}

LLMs are demonstrating increasingly sophisticated cognitive abilities, including idea generation~\cite{ege2024chatgpt}, problem-solving~\cite{rasal2024optimal}, and the potential to simulate believable human behaviors such as reflection and planning~\cite{park2023generative}. These capabilities are enabling LLMs to assist significantly in professional workflows and data-driven decision-making across various sectors, notably in Law~\cite{siino2025exploring} and Medicine~\cite{kim2024mdagents}. For example, LLMs are already assisting with tasks such as drafting documents, reviewing information, and generally speeding up rote, repetitive, or data-intensive tasks within the decision making processes that are prone to human error or consume significant time~\cite{guha2023legalbench,ma2024towards}. However, despite this demonstrated potential and application in other professional domains, their widespread adoption in government decision-making processes remains limited, primarily due to concerns surrounding trust and reliability.

In this work, we assess \gls*{llm} performance in the domain of environmental reviews conducted under the National Environment Policy Act (\gls*{nepa}\footnote{https://www.epa.gov/nepa}).
% Environmental Impact Statements (\gls*{eis}) for major federal actions are comprehensive documents detailing proposed actions, their alternatives, and potential environmental impacts, while demonstrating compliance with environmental laws and executive orders.
\gls*{nepa} stands as a foundational piece of environmental legislation in the United States, requiring federal agencies to consider the environmental impacts of their proposed actions. Beyond simply protecting the environment, the NEPA process plays a important role in promoting societal progress and responsible economic growth. By assuring informed decision-making, promoting transparency, and incorporating public input, NEPA helps shape federal decisions that are more sustainable, resilient, and responsive to community needs. This proactive approach can prevent costly environmental impacts, facilitate the development of necessary infrastructure in an environmentally sound manner, and build public trust, ultimately contributing to long-term economic stability and community well-being.

However, the NEPA decision making process is inherently complex, and characterized by different agency-wide regulations, multi-disciplinary technical analysis, and a deeply collaborative process involving diverse stakeholders including applicants, consultants, regulatory agencies, legal teams, and the public\footnote{https://www.epa.gov/nepa/national-environmental-policy-act-review-process}. 
To perform such decisions, it requires more than just comprehensive knowledge of federal, state, and local laws; it demands the ability to interpret their application to specific site conditions, analyze complex environmental interactions, predict potential outcomes, and communicate findings clearly and effectively. NEPA decision makers want the LLMs to perform the higher-order regulatory focused reasoning beyond fact retrieval for effective decision support. To support the adoption of LLMs in NEPA decision making workflows\footnote{https://www.whitehouse.gov/presidential-actions/2025/04/updating-permitting-technology-for-the-21st-century/}, we test the LLM's abilities to perform regulatory focused reasoning to interpret the intent behind regulations, apply their principles effectively, and support logical deductions or evaluations grounded in NEPA regulatory frameworks. This involves evaluating potential compliance pathways, identifying likely environmental impacts based on regulatory criteria, recommending appropriate mitigation strategies, and predicting plausible agency review outcomes based on project characteristics.

\begin{figure*}[!t]
\centering
    \includegraphics[width=\linewidth]{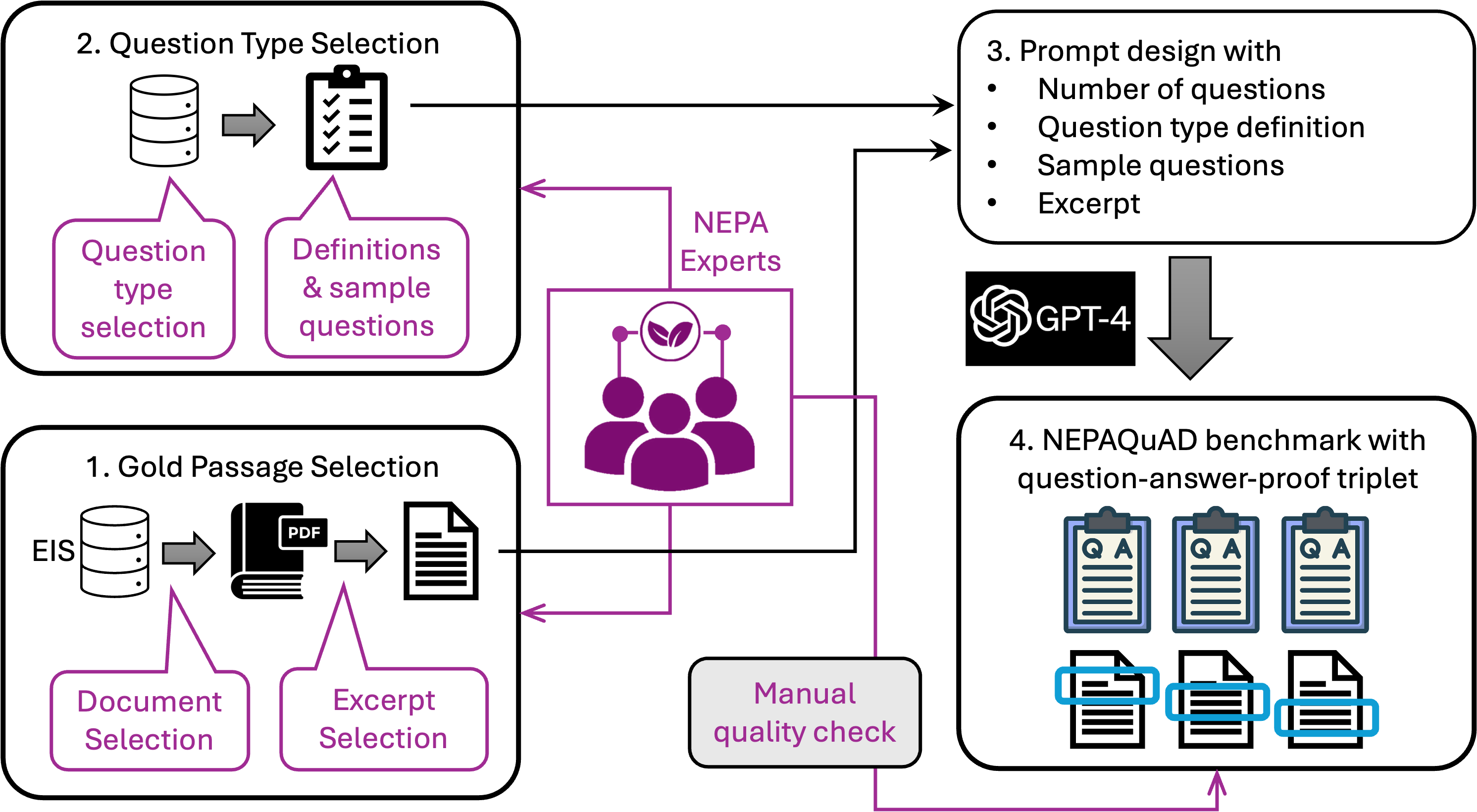}
\caption[short]{Steps of Ground-truth benchmark generation for evaluating \gls*{llm}s over varied contexts for question-answering over EIS documents}
\label{fig:diagram:BenchmarkGeneration}
\end{figure*}

% While creating domain-specific benchmarks like \nepaquad is crucial, their effective utilization requires a standardized evaluation framework. 
% This challenge becomes increasingly complex with multiple frontier \gls*{llm}s hosted by different providers, varied context types and text generation prompts, and different benchmark formats.
% To address this gap, we developed Multi-context Assessment Pipeline for Language model Evaluation (MAPLE), a modular evaluation pipeline that seamlessly handles response generation across \gls*{llm} providers. 
% \maple also supports various evaluation scenarios -- basic question-answering with no context, and advanced features like \gls*{rag}-based response generation and context-grounded evaluation. 
% Unlike existing frameworks such as RAGAS~\cite{ragas} and DeepEval~\cite{deepeval}, MAPLE provides a comprehensive solution specifically designed for large-scale assessments involving multiple models and context types. 
% Where RAGAS focuses primarily on metric definitions and DeepEval on detailed test configurations with a particular focus on OpenAI models, MAPLE streamlines the entire evaluation process with unified provider interfaces, built-in progress management, and automatic context type adaptation, allowing researchers to evaluate existing benchmarks across different models with minimal setup overhead.
% \gls*{nepa} is a U.S. environmental law designed to protect the environment. 

An environmental impact statement (EIS) is required by Section 102(2)(C) of \gls*{nepa} for any proposed major federal action significantly affecting the quality of the environment. An EIS is a detailed document prepared that describes a proposed action, alternatives to the proposed action, and potential effects of the proposed action and alternatives on the environment. An EIS contains information about environmental permitting and policy decisions and considers a range of reasonable alternatives, analyzes the potential impacts resulting from the proposed action and alternatives, and demonstrates compliance with other applicable environmental laws and executive orders. Along with the fact that EIS documents are usually lengthy (often several hundred pages) and are created by \gls*{nepa} experts, another factor that can hinder the application of \gls*{llm}s in this domain is that the development of an EIS document requires \gls*{nepa} experts with various subject matter expertise to engage in preparation over multiple years, often citing older articles from as far back as the 1990s. For example, the Executive Order (EO) 12898, issued in 1994, is cited on page 60 of the EIS documents prepared for the First Responder Network Authority project\footnote{https://www.energy.gov/nepa/eis-0530-nationwide-public-safety-broadband-network-programmatic-environmental-impact}. This could present significant challenges for current \gls*{llm}s in helping \gls*{nepa} users automatically retrieve answers from \gls*{llm}-based question-answering systems. To our knowledge, there is no ground-truth benchmark built specifically for this domain to evaluate the quality of \gls*{llm}s' output for QA task when the questions pertain to EIS documents.

To this end, we evaluate the extent to which \gls*{llm}s can comprehend \gls*{nepa} knowledge for diverse question typologies specifically designed to assess NEPA focused regulatory reasoning over \gls*{eis}. %(Figure~\ref{fig:diagram:ContextStudy}). 
We select frontier \gls*{llm}s for our experiments: Claude Sonnet \cite{ClaudeSonnet}, Gemini-1.5Pro \cite{geminiteam2024gemini}, GPT-4 \cite{openai2024gpt4}, Llama 3.1\cite{dubey2024llama} and Mistral-7B-Instruct\cite{jiang2023mistral}. 
We built \textbf{NEPA} \textbf{Qu}estion and \textbf{A}nswering \textbf{D}ataset (\textit{NEPAQuAD}) benchmark that includes the triplets of questions, answers, and corresponding contexts, generated through a hybrid approach employing GPT-4 and NEPA experts. 
To support our evaluation, we developed a \textbf{M}ulti-context \textbf{A}ssessment \textbf{P}ipeline for \textbf{L}anguage model \textbf{E}valuation (MAPLE), a transparent and modular evaluation pipeline that seamlessly handles response generation across different \gls*{llm} providers. 
\maple also supports various evaluation scenarios -- basic question-answering with no context, and advanced features like \gls*{rag}-based response generation and context-grounded evaluation.
Using the \maple framework, we conduct rigorous experiments evaluating frontier \gls*{llm}s with various types of contexts for \gls*{nepa} documents. 
Overall, we make the following contributions:
\vspace{-0.25em}
\begin{enumerate}
    \item Constructed the first-ever benchmark \nepaquad to evaluate the performance of \gls*{llm}s for diverse question typologies specifically designed to assess NEPA focused regulatory reasoning.
    \item Developed \maple as a standardized evaluation pipeline to compare the capabilities of \gls*{llm}s in various context-driven prompting strategies (i.e., gold passage, entire PDF document, and \gls*{rag}) to assess model performance.
    \item NEPA experts driven performance evaluation in the \gls*{llm}s ability to support real-world regulatory decisions.
\end{enumerate}
\vspace{-0.25em}
The structure of this paper is outlined as follows. In Section~\ref{sec:benchmark}, we describe the \nepaquad benchmark in detail. Section~\ref{sec:maple} lays out our approach and the various contexts used for evaluation implementation, followed by a detailed analysis of our performance in Section~\ref{sec:results}. Section~\ref{sec:related_work} then discusses other socio-scientific benchmarks and evaluation frameworks in the literature. We finish with the conclusion and limitations of our work in Sections~\ref{sec:conclusion} and \ref{sec:limitations}.
\section{The \nepaquad Benchmark}
\label{sec:benchmark}

We present \nepaquad (\textbf{N}ational \textbf{E}nvironmental \textbf{P}olicy \textbf{A}ct \textbf{Qu}estion and \textbf{A}nswering \textbf{D}ataset Version 1.0), the first-ever  benchmark to evaluate the performance of LLMs in a question-answering task for EIS documents. NEPAQuAD consists of 1590 questions, split into two different question types: open and closed. The open questions are further split into nine different types, as shown in Table \ref{tbl:qt:distribution}. Overall, these questions not only require \gls*{llm}s to process and comprehend long-context documents, but also reason in the field of environmental permitting. As such, these questions are perfectly suited to serve as the test for any \gls*{llm}'s capability to reason in this regulatory domain.

% To evaluate performance of \gls*{llm}s for the EIS question-answering task, we first select high-quality documents from the EIS document database and extract paragraphs as context to be used in the evaluation. Then, we identify the types of questions that we want to use to evaluate the models. Next, we use GPT-4~\cite{openai2024gpt4} to generate question-answer pairs based on the selected contexts by using carefully designed prompts. Finally, we use these generated questions to evaluate different \gls*{llm}s with various contexts, with the generated answers acting as the ground-truth. We describe the process in detail below.

\subsection{Dataset Creation}

Due to the high costs associated with manually creating the entire dataset and the inability to use ground-truth benchmarks from other domains, we adopt a hybrid approach with GPT-4 and NEPA experts to generate the \nepaquad benchmark. The process we followed for this benchmark generation is illustrated in Figure \ref{fig:diagram:BenchmarkGeneration}, which we describe in more detail in the following sections:

% The general idea of our evaluation benchmark generation process is to extract meaningful passages from a set of \gls*{eis} documents, then use GPT-4 to generate questions based on these passages. To ensure the quality of the generated benchmark, two authors of this study, who are subject matter experts in \gls*{nepa}, measure the quality of the ground-truth answers by comparing the provided proofs against the original context from which the questions were derived. Our generated ground-truth benchmark is a triplet containing a question, the answer to the question, and the proof (i.e., the text directly related to the answer, derived from the context from which the question originated). 

\noindent \textbf{Document Curation}
The first task is to curate a diverse set of high-quality \gls*{eis} documents from federal databases and select excerpts to establish a representative evaluation corpus. \gls*{nepa} experts selected nine \gls*{eis} documents from different government agencies that were most representative of various \gls*{nepa} actions. These documents exhibit significant variation in content and focus depending on the authoring government agency, as each agency may interpret and implement the \gls*{nepa} guidelines distinctively. For instance, the U.S. Forest Service might emphasize forest management and wildfire mitigation, while the U.S. Army Corps of Engineers could prioritize water resource development and infrastructure impacts. Please refer to Table~\ref{tbl:dataset:statistics} in the Appendix for a brief overview about the selected documents. We consider documents of varied lengths, with the longest document containing nearly 900 pages (>600K tokens).

\noindent \textbf{Gold Passage Selection} For each of the nine selected documents, we attempt to select excerpts that have important content of each document. A naive approach of randomly extracting passages poses the risk of resulting in \emph{low-quality} excerpts, such as parts of appendices or image captions. To avoid this risk, \gls*{nepa} experts manually select excerpts from the documents. 
They divided each document into three sections -- beginning, middle, and end, and then selected two, six, and two excerpts from each of these sections respectively, resulting a total of 10 excerpts from each document. We use these excerpts as the ground-truth context, called \emph{gold passages}, for question benchmark generation. 

\noindent \textbf{Question Type Selection}
% Once we identified the gold passages, \gls*{nepa} experts select the type of questions to include in the benchmark. We started with a list of 15 types of questions\footnote{https://tinyurl.com/3akej8ct}, and eventually narrowed it down to 10 types of questions 
% \sameera{Please add two sentences demonstrating how diverse the questions}
% The benchmark encompasses a wide spectrum of questions, challenging the \gls*{llm}s with everything from straightforward `closed' questions requiring binary responses to complex `problem-solving' scenarios demanding insightful responses. 
% The cognitive diversity of the questions forces the models to demonstrate both precise procedural knowledge as well as reasoning ability.
Next, we developed a question taxonomy that encompasses various complexities inherent in environmental impact assessments. After extensive discussions with \gls*{nepa} experts, we finalized 10 types of questions that facilitate the NEPA decision making process. These question types test the LLM's ability to articulate the purpose of regulations, explain complex review processes, or describe the steps involved in a permit application.

For example, \textit{funnel} questions regarding the environmental alternative analysis, such as \textit{`Which alternatives were discussed?', `Which were considered?' `Why was [ALTERNATIVE] not considered?'} require in-depth environmental and legal knowledge to justify these decisions. On the other hand, problem solving questions such as \textit{`Given the location of the [PROJECT], create a list of aquatic species likely present in a 50-mile radius'} present specific regulatory scenarios to test the ability of the models to comprehend the latest geographical knowledge. Overall, we ensure almost half the questions in the dataset are these types of `open' questions. We present the question types and relevant examples in Table \ref{tbl:qt:distribution}.

% \anurag{Let's add some sample questions here.}
% \rounak{added Figure 2}
\begin{figure*}[!t]
\centering
    \includegraphics[width=\linewidth]{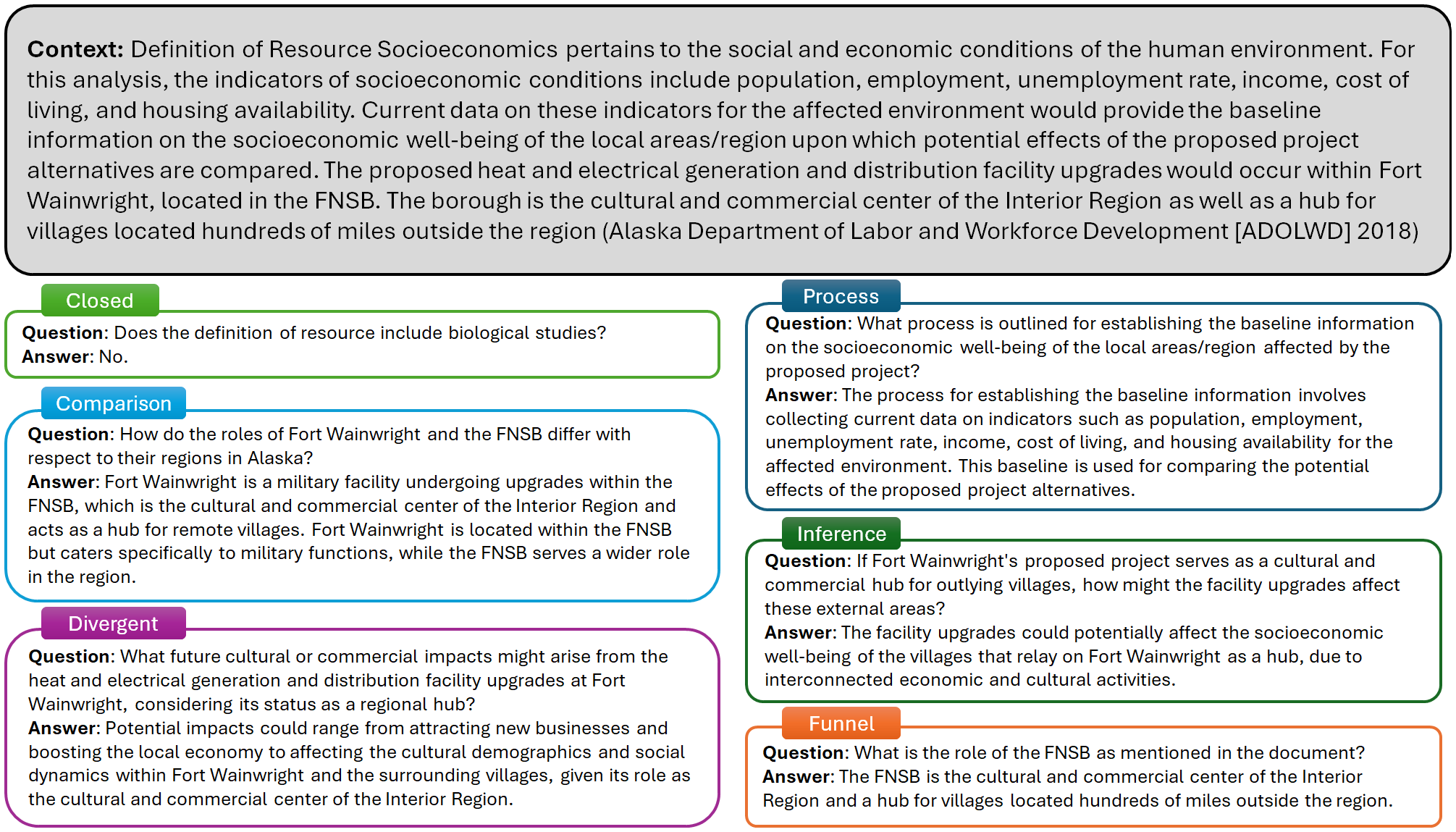}
\caption[short]{Examples of question-answer pairs present in \nepaquad, and the context used to generate these questions.}
\label{fig:sampleQA}
\end{figure*}

In addition to selecting the question types, the \gls*{nepa} experts also created sample questions for each type. For a more detailed description of the question types, as well as example questions provided by the \gls*{nepa} experts, please refer to Appendices~\ref{sec:question_defs} and \ref{sec:question_samples}.

\noindent \textbf{Prompt Design}
We leverage generative models (such as GPT-4) with carefully crafted prompts to generate question-answer-proof triplets based on the selected \gls*{eis} contexts, ensuring coverage across different question types. To ensure that the prompts can instruct generative models efficiently, we consulted with the \gls*{nepa} experts to create the prompts. We also use the sample questions created by the \gls*{nepa} experts to augment the original prompts and create an \emph{enhanced} prompt. The template for the prompt and benchmark creation process is provided in Appendix \ref{app:prompt}.

\begin{table}
\centering
\caption{Question types in the \nepaquad benchmark}
\begin{tabular}{lrrlrr}
\hline
\multicolumn{1}{l}{\textbf{Type}} & \multicolumn{2}{r}{\textbf{\#Questions}} & \multicolumn{1}{l}{\textbf{Type}} & \multicolumn{2}{r}{\textbf{\#Questions}}                      \\ \hline
Closed                                       & 789         & 49\%                                &
Comparison                                   & 64            & 4\%                                 \\ 
Convergent                                   & 109          & 7\%                                 &
Divergent                                    & 121           & 8\%                                 \\ 
Evaluation                                   & 64          & 4\%                                 &
Funnel                                       & 127          & 8\%                                 \\ 
Inference                                    & 101         & 6\%                                 &
Problem-solving                              & 11          & 1\%                                 \\ 
Process                                      & 108         & 7\%                                 &
Recall                                       & 105         & 7\% \\ \hline
\end{tabular}
\label{tbl:qt:distribution}
\end{table}

\noindent \textbf{Automated Generation} Following successful precedents in other domain specific benchmark creation~\cite{balaguer2024rag}, we employed GPT-4 through Azure OpenAI service with default parameters to generate the question-answer-proof benchmark triplets.
Our dataset construction began with nine carefully selected \gls*{eis} documents, from which we extracted 10 gold passages each, totaling 90 passages. For each passage, we generated a diverse set of questions: 10 closed-type questions and two questions each from nine other question categories. While the initial generation yielded 2520 questions, we implemented a rigorous quality control process. First, we filtered out responses that did not conform to our specified output format, reducing the count to 1920.
Subsequently, we conducted manual quality assessment of the remaining questions, resulting in a final benchmark of 1590 high quality question-answer-proof triplets across the selected \gls*{eis} documents. 
Few examples of question-answer pairs generated for a gold passage context is illustrated in Figure~\ref{fig:sampleQA}.
\nepaquad is made publicly available~\citep{llm-for-environmental-review} to facilitate further research and development in this domain.

\subsection{Quality Assessment}

\noindent \textbf{Comparison with other QA datasets.}~
We test the intrinsic quality of the generated questions with several metrics related to content and style, and compare the metrics against the similar QA datasets SQuAD~\cite{rajpurkar2016squad} and SciQ~\cite{welbl2017crowdsourcing}.
The following metrics are used to assess data quality:
\begin{enumerate}
    \item Entailment between the context and the question and answer, measured via prediction from a BERT model fine-tuned on natural language inference (NLI)\footnote{Accessed 1 April 2025: \url{https://huggingface.co/tasksource/ModernBERT-base-nli}};
    \item Complexity of the question, measured via parse tree depth from a spacy dependency parser (\texttt{en\_core\_web\_trf} model);
    \item Specificity of the question, measured via number of named entities extracted using a SpaCy\footnote{https://spacy.io/} named entity recognition model.% (\texttt{en\_core\_web\_trf} model).
    \item Readability of the question, measured via Flesch-Kincaid readability score~\cite{kincaid1975derivation}.
\end{enumerate}

\begin{table}%[!h]
    \centering
    \caption{Comparing NEPAQuAD with existing QA benchmarks using various question quality metrics}
    \begin{tabular}{l r r r} \toprule
        Metric & NEPAQuAD & SQuAD & SciQ \\ \midrule
        \% Entailment + Neutral & 0.925 & 0.985 & 0.968 \\
        Complexity & 5.317 & 3.867 & 4.741 \\
        Specificity & 1.17 & 1.11 & 0.139 \\
        Readability & 34.4 & 60.5 & 58.0 \\ \bottomrule
    \end{tabular}
    \label{tbl:quality_metrics}
\end{table}

The results in Table \ref{tbl:quality_metrics} show that \nepaquad is on-par or exceeding in the metrics as compared to similar QA datasets. For instance, the questions in \nepaquad are more complex than SQuAD and SciQ, while the proportion of questions with entailment or neutral NLI labels is in a similar range to the other datasets. The lower readability score for \nepaquad indicates longer sentences and a more complex vocabulary, which is reasonable considering the NEPA domain.

\begin{table*}%[!ht]
\centering
\caption{Summary of annotation agreement among three annotators.}
\label{tab:agreement_summary}
\begin{tabular}{llll}
\hline
\textbf{Aspect} & \textbf{\% Majority Yes} & \textbf{\% Majority No} & \textbf{\% Unanimous Agreement} \\ \hline
Is question type correct? & 81.0 & 19.0 & 89.0 \\
Is the answer correct? & 93.0 & 7.0 & 71.0 \\
Is the proof correct? & 93.0 & 7.0 & 79.0 \\ \hline
\end{tabular}
\label{tbl:majority-stat}
\end{table*}

\begin{table*}[!ht]
\centering
\caption{Annotator agreement scores with ground truth (majority rule)}
\label{tab:annotate_agreement}
\begin{tabular}{cllll}
\hline
\textbf{} & \textbf{} & \textbf{Is question type correct?} & \textbf{Is the answer correct?} & \textbf{Is the proof correct?} \\
 \hline
\multirow{2}{*}{\begin{tabular}[c]{@{}c@{}}\gls*{nepa} \\ expert 1\end{tabular}}  & \% Agreement   & 99.0           & 81.0     & 87.0     \\
 & Cohen's Kappa  & 0.967 (Almost Perfect)          & 0.328 (Fair)    & 0.340 (Fair)    \\
 % & Interpretation & Almost perfect & Fair     & Fair     \\ 
 \hline
\multirow{2}{*}{\begin{tabular}[c]{@{}c@{}}\gls*{nepa}\\ expert 2\end{tabular}}  & \% Agreement   & 90.0           & 90.0     & 92.0     \\
 & Cohen's Kappa  & 0.734 (Substantial)         & 0.423 (Moderate)    & 0.560 (Moderate)    \\
 % & Interpretation & Substantial    & Moderate & Moderate \\ 
 \hline
\end{tabular}
\end{table*}

\noindent \textbf{Human Review} To ensure a high-quality benchmark, we employed a rigorous annotation protocol. We randomly sampled 100 benchmark entries from the entire benchmark dataset. Two independent \gls*{nepa} experts evaluated each sampled entry across three critical dimensions: the correctness of question type (i.e. whether the generated question was the same type of question as requested), the correctness of the generated answer, and the correctness of generated proof. For each aspect, annotators provided binary judgments (`yes'/`no') based on established criteria documented in the annotation guidelines provided to the experts. For entries where the \gls*{nepa} experts disagreed on one or more aspects, we engaged a third independent \gls*{nepa} expert to adjudicate, creating a three-annotator subset. To determine overall quality and establish ground truth, we then employed a majority voting approach. These scores are shown in Table~\ref{tab:agreement_summary}. Furthermore, we evaluated each annotator's alignment with this majority opinion by calculating individual Cohen's Kappa scores between each expert's responses and the majority vote as shown in Table~\ref{tab:annotate_agreement}. We show examples of human annotation of the generated triplet in Appendix \ref{sec:human-annotation-example}. 

\section{The \maple Evaluation framework}
\label{sec:maple}

We developed \maple (Multi-context Assessment Pipeline for Language model Evaluation)\footnote{\url{https://github.com/pnnl/permitai/evaluation}} as a comprehensive evaluation tool that streamlines the assessment of \gls*{llm}s in question answering and document retrieval tasks. As illustrated in Figure~\ref{fig:pipeline}, we designed it to provide a unified interface supporting multiple \gls*{llm} providers while ensuring consistent evaluation methodologies. We incorporated a modular design that we used to integrate various context types -- from no-context baselines to \gls*{rag}-enhanced setups. We also implemented RAGAS evaluation metrics~\cite{es2023ragas}, allowing us to benchmark \gls*{llm} performance across different benchmarks. {The \maple architecture comprises four modules: a flexible \emph{dataloader} that standardizes benchmark entries from various file formats, a provider-agnostic \emph{\gls*{llm} handler} enabling unified interaction across multiple \gls*{llm} services, an \emph{evaluator} module that dynamically populates prompt templates with questions and context information from benchmark entries, and a comprehensive \emph{metrics} module supporting both user-defined custom metrics and established metric evaluation frameworks like RAGAS~\cite{ragas}. We discuss the key features of \maple below. Please refer to Appendix~\ref{ref:app-sec-maple} for a detailed overview of individual modules.
% \sameera{I see that you describe individual modules in the sections later, you have to introduce these modules in briefly here.}

\begin{figure}
    \centering
    \includegraphics[width=0.47\textwidth]{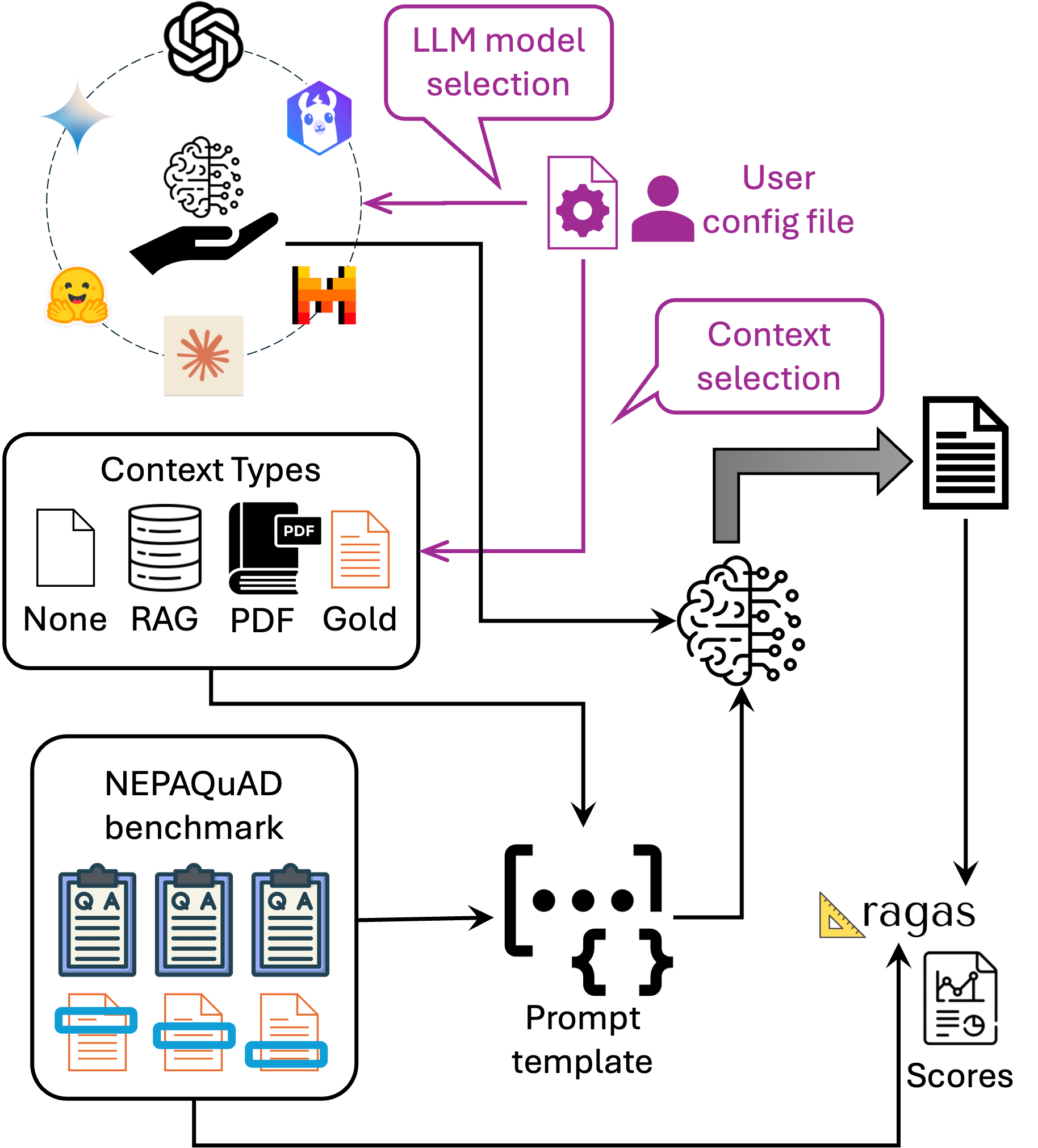}
    \caption{Overview of the MAPLE pipeline.
    % \sameera{can we change the Figure 2 to look consistent with Figure 1 formatting and styles?}
    }
    \label{fig:pipeline}
\end{figure}

\subsection{Supporting multiple LLM providers}
We implemented an \textit{\gls*{llm} Handler} module using an abstract base class to provide support for major cloud services. 
Our implementation includes Azure OpenAI (GPT models), AWS Bedrock (Claude, LLama, Mistral), Google Vertex AI (Gemini), and local HuggingFace models. 
We designed the handler to manage provider-specific authentication requirements and parameter specifications while maintaining a standardized interface for response generation.
In this way, we effectively abstract the complexities of working with different \gls*{llm} services into a single, consistent evaluation pipeline.

% However, we included this triplet in \nepaquad since at least one of the evaluators accepted the generated triplet.

\subsection{Supporting multiple context types.}
\label{ssec:contexts}
The \textit{Evaluator} module (which generates \gls*{llm} responses to the questions in the benchmark) supports four context types to enable comprehensive evaluation across different scenarios. These include a basic \textit{no-context} mode to test prior knowledge of \gls*{llm}, a \textit{PDF} mode for processing complete documents, a \textit{RAG}~\cite{lewis2020retrieval} mode for vector database integration, and a \textit{gold} context mode for controlled evaluations with the gold passage as context. \maple automatically identifies and validates appropriate context types by analyzing the available fields in benchmark entries. It defaults to the `no-context' mode when only question-answer pairs are present.
% \sameera{Following 4 sentences are not clear, they may represent some redundant information, please simplify and improve.}
% We designed the framework to automatically detect and validate appropriate context types based on the benchmark dataset structure.
% Using the different fields (for example, columns in the CSV file of \nepaquad benchmark), we identify the possible context types which can be supported by \maple.  
% Otherwise, \maple automatically defaults to the no-context mode when only question-answer pairs are present. 
% Through this automatic detection and validation system, we ensured consistent and appropriate context handling across different evaluation scenarios.
Here, we provide brief description of these four context types.

% \begin{figure}[h]
% \centering
%     \includegraphics[width=\linewidth]{images/diagrams/FlowDiagram.png}
% \caption[short]{A simple illustration of varied contexts used in the comparison. \anurag{I don't see any value of this figure.}}
% \label{fig:diagram:ContextStudy}
% \end{figure}

\noindent\textbf{No Context:} Questions are presented to the models without any additional context. As such, one can think of this mode as analogous to using an LLM-based chatbot. While this strategy can work well with popular domains such as sport or literature, we assume that the \gls*{nepa} domain may be challenging for the \gls*{llm}s to provide accurate answers. In other words, for our benchmark, this setting can be considered a test of the \gls*{llm}s' ability to answer out-of-general-domain questions. As such, this setting is usually expected to return low performance.

\noindent\textbf{PDF Context:} In addition to the question, we also provide the model the text from the entire PDF document from which the question was generated. Since we do not inform the model which part of the document to look at, the accuracy of generated responses will heavily depend on the models' ability to pick the correct context from the very large scale textual information provided. We expect this setting to yield performance better than the \textit{no context} scenario. This setup evaluates the models' ability to process and extract relevant information from lengthy, complex documents.

\noindent\textbf{\gls*{rag} Context:}~
% \note{is "Augmented Context" a better word choice here?}
Employing a \gls*{rag} approach, this setup provides the top-$k$ most relevant text chunks retrieved from a vector database containing text chunks of \gls*{nepa} documents. In our \gls*{rag} model implementation, when a question is inputted for \gls*{llm} generation, the corresponding context is extracted as a relevant passage from a given \gls*{nepa} document. We use the standard \gls*{rag} setup where we encode both question and retrieved passages with the BGE embedding model~\cite{bge_embedding}. Cosine similarity scores are used to assess the similarity between the question and the contexts. In our case, the number of top-ranked relevant text chunks extracted is set at $k=3$. This scenario assesses the models' performance when given highly relevant, focused information that is specifically retrieved to answer the question at hand.

\noindent\textbf{Gold Context:}~
In this configuration, we include the actual text excerpt from which the question was generated in the prompt, alongside the question content. This scenario represents an ideal case where the most relevant information is perfectly identified and provided. While the situation where users manually identify the correct text passage is rare in practice, we simulate this scenario to measure how well \gls*{llm}s can perform if we were able to extract relevant passages with very high accuracy. This setup serves as an upper bound for performance, testing the models' ability to comprehend and utilize perfectly relevant information.

\subsection{Supporting custom prompt templates}
We designed the prompt engineering system of \maple to be flexible and customizable through a template-based approach. We implemented a system where users can provide prompt templates through external files, with designated placeholders for \textit{question} and \textit{context} elements. The system automatically formats these templates during evaluation by replacing placeholders with actual content. We also included validation checks to ensure required placeholders are present for specific context types. For example, we require both question and context placeholders for \textit{gold} context mode evaluations, while basic \textit{no-context} evaluation requires only the placeholder for question.  An example prompt template with placeholders for question and context is shown below. Through this design, we enabled users to experiment with different prompting strategies while ensuring consistent prompt structures across evaluations.

\prompt{You are a helpful assistant who will answer questions about environmental policies from a context provided in the prompt.\\
Question: \{placeholder for question\}\\
Answer the question from the context.\\
Context: \{placeholder for context\}}

\subsection{Evaluation metrics}
\label{ssec:eval_metrics}
We leveraged the RAGAS evaluation framework~\cite{es2023ragas} to assess \gls*{llm} responses against benchmark ground truth answers. 
We specifically focused on the \emph{answer correctness} score, which is configured to combine two key components: factual correctness and semantic correctness. 
For factual correctness, which comprises $75\%$ of the final score, we utilized GPT-4 to evaluate the accuracy of generated answers at the phrase/clause level compared to ground truth answer in the benchmark. 
We complemented this with semantic correctness, weighted at $25\%$, where we employed the BGE embedding model~\cite{bge_embedding} to compare vector embeddings of predicted and expected answers. 
We chose this comprehensive evaluation approach over traditional overlap-based metrics like BLEU scores\cite{036_BLEU}, as it efficiently captures both factual accuracy and semantic similarity of the generated responses and ground truth answers.

We employ a separate evaluation method for closed type questions which have only `yes' or `no' answers using the following steps. First, we process the responses generated by the \gls*{llm}s to extract only the `yes' or `no' answer from the generated response. Any additional explanatory text or elaboration is removed from the response. Thereafter, the cleaned binary response is compared directly with the ground truth answer. If the cleaned response exactly matches the ground truth, it is assigned a full score of $100\%$. If there is any discrepancy, including cases where the model fails to provide a clear "Yes" or "No" answer, the response is scored as $0\%$. This straightforward evaluation method for closed questions allows us to assess the models' ability to provide accurate binary responses without being influenced by any uncertainties introduced by the RAGAS evaluation framework. We refer this metric as \textit{answer correctness} score for closed type questions.
\section{Performance Analysis}
\label{sec:results}
In this section, 
% we detail the combined usage of \nepaquad and \maple. 
we examine the performance of the \gls{llm}s on \nepaquad using the \maple framework (as presented in Section~\ref{sec:maple}). 
First, we compare the performance of the five frontier \gls{llm}s%: GPT-4, Claude-3 Sonnet, Gemini, Mistral and Llama3.1 
across various contexts (Section~\ref{ssec:eval_qa_contexts}). Then, we compare the model performance across various question types (Section~\ref{ssec:eval_qa_types}). %Finally, we analyze the performance.

\subsection{Evaluating Different QA Contexts}
\label{ssec:eval_qa_contexts}
Table~\ref{policyai:tbl:scores} reports the overall performance of the models across various QA contexts used in the evaluation. We observed that for the task with \textit{no context}, the GPT-4 model produces the most accurate results by far. However, when PDF documents are provided as context, Claude surpassing GPT-4 and Gemini in the answer correctness. Despite the fact that Gemini is able to handle very long contexts (1.5M tokens), it is surprising to see its performance drop when provided with PDF documents as additional contexts. This may be due to the model struggling to reason over the large amount of relevant and irrelevant content in the EIS document. Open-source models (LLama3.1 and Mistral) generally underperform compared to closed \gls{llm}s across most context types, including PDF, \gls{rag}, and gold contexts. However, in the no-context scenario, all models exhibit similar performance, suggesting that the availability and quality of context significantly influence the performance gap between open-source and closed models.

Overall, \gls{rag} models perform better in comparison to the models provided with PDF documents as additional contexts. In \gls{rag} setup, the Gemini model outperforms other models in term of correctness, although the correctness scores of Claude and Gemini models are much closer. There is notable increase in Llama3.1's performance in the \gls{rag} setup when compared with the PDF contexts. This suggests that the performance of long-context models can be improved when provided with the most relevant context, as in the \gls{rag} setup.

As expected, all models perform best on average when provided with the gold passage in comparison to other context variations, as in this scenario, models synthesize information that directly contains the answer to the question posed to the model. Notably, models perform comparably when they are provided with the \gls{rag} and gold passage contexts. The only exception to this trend is the Llama3.1 model, which outperforms in the \gls{rag} setup as compared to the gold passage.

\begin{table}
\centering
\caption{Answer correctness of \gls{llm} responses on the \nepaquad benchmark across different context types. The best-performing setting for each model is shown in bold.}
\label{policyai:tbl:scores}
\begin{tabular}{|l|l|l|l|l|}
\hline
\textbf{Context} & None & PDF & RAG & Gold  \\ \hline
\textbf{GPT-4} & 67.00\% & 63.70\% & 74.36\% & \textbf{76.60}\%\\ \hline
\textbf{Claude} & 64.53\% & 66.46\% & 75.16\% & \textbf{76.84}\%\\ \hline
\textbf{Gemini} & 62.84\% & 65.90\% & 75.46\% & \textbf{81.15}\% \\ \hline
\textbf{Mistral} & 64.95\% & 61.81\% & 72.88\% & \textbf{75.34}\% \\ \hline
\textbf{Llama3.1} & 66.35\% & 59.52\% & \textbf{74.01}\% & 72.73\%\\ \hline
\end{tabular}
\end{table}

\subsection{Evaluating Different Question Types}
\label{ssec:eval_qa_types}
\begin{figure*}[ht]
\centering
\includegraphics[width=0.32\textwidth]{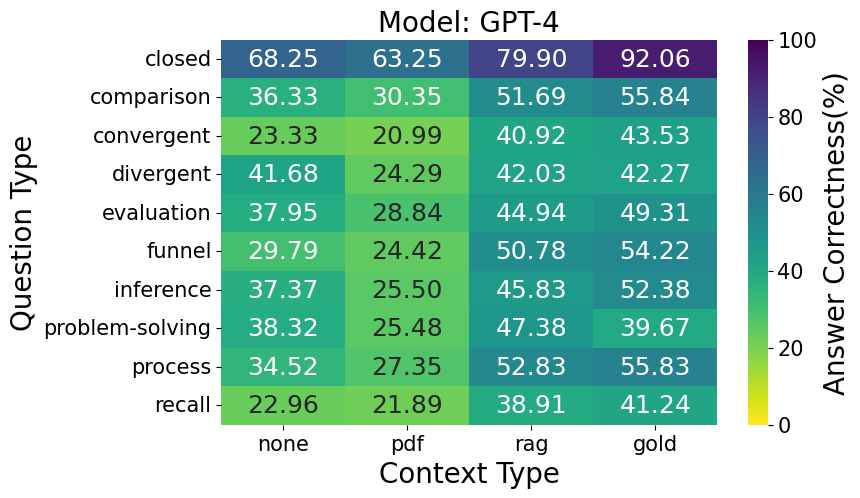}
\includegraphics[width=0.32\textwidth]{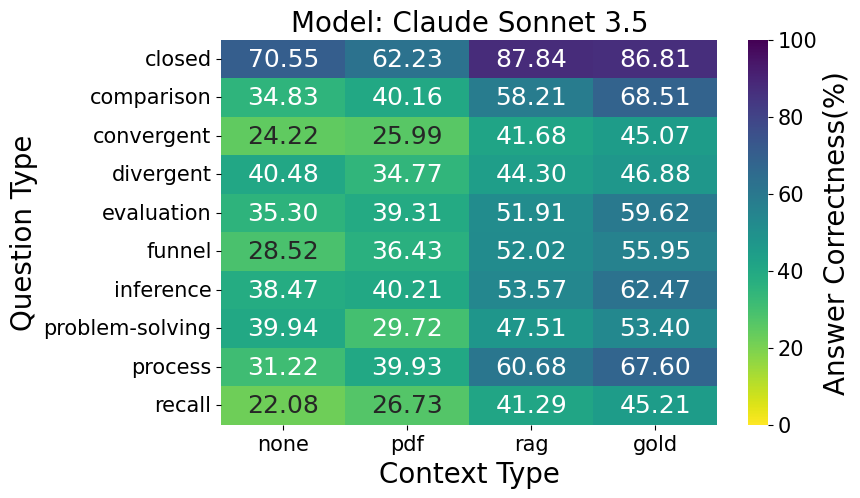}
\includegraphics[width=0.32\textwidth]{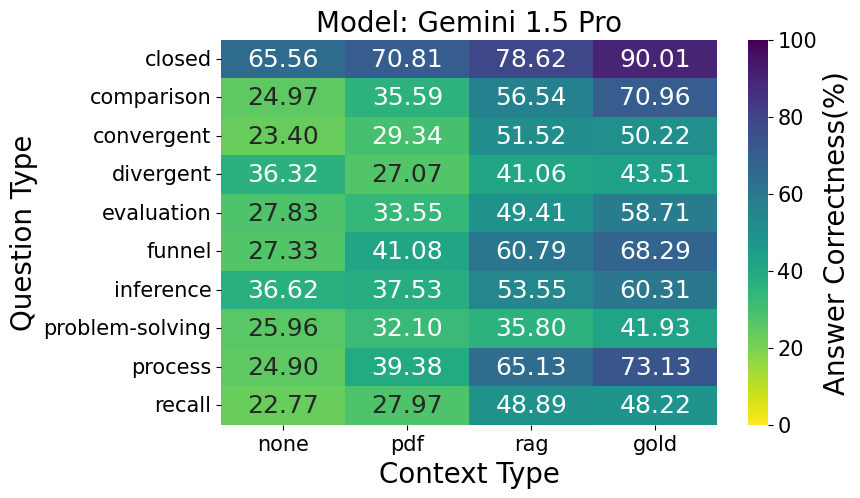}
\includegraphics[width=0.32\textwidth]{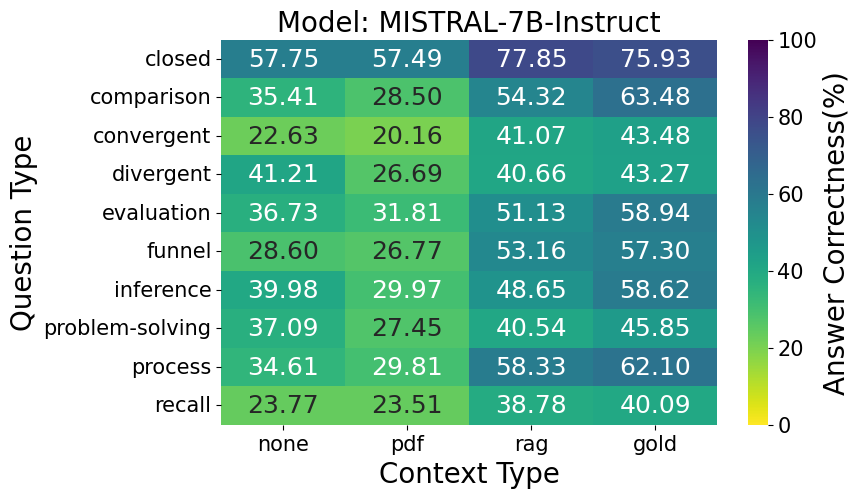}
\includegraphics[width=0.32\textwidth]{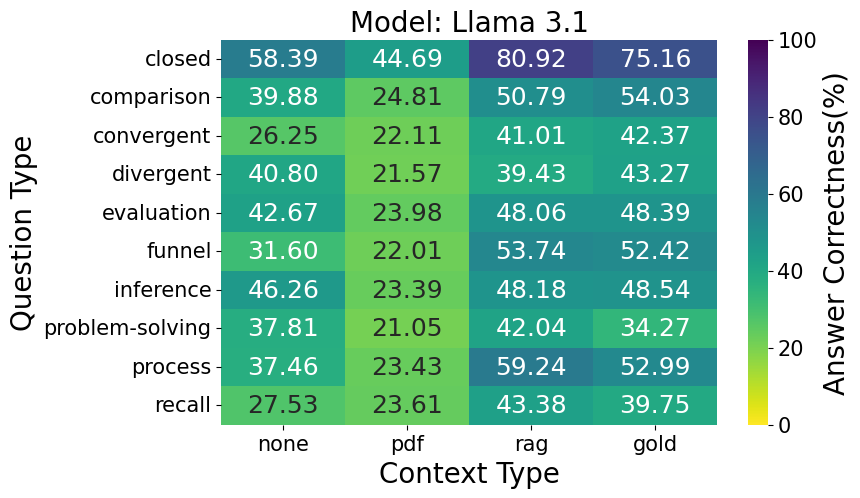}
\caption[short]{The evaluation results meassured by the Answer Correctness scores of each LLM used with 4 scenarios of using context over each question types}
\label{fig:accuracy:questionTypes}
\end{figure*}

Figure \ref{fig:accuracy:questionTypes} shows the performance of the models across different question types. The \textit{process}, \textit{comparison} and \textit{evaluation} questions generally benefit from richer contexts, with closed-source models demonstrating a more significant improvement. \textit{Convergent} and \textit{recall} type questions prove challenging for all models, especially in no-context scenarios, highlighting the difficulty of synthesizing information or retrieving specific details without supporting context. \textit{Problem-solving} questions show varied results across models, with some struggling even with rich context, indicating that complex analytical tasks remain a challenge. \textit{Funnel} questions benefit significantly from context, especially for models like Gemini 1.5, suggesting that structured, narrowing inquiries are well-suited to context-enhanced responses. The \textit{closed} questions yield the highest accuracy for all models, regardless of context. It is important to note that the evaluation metric for closed questions directly compares the model's \textit{yes}/\textit{no} response to the ground truth, in contrast to the RAGAS metric used for other question types.
\section{Related Works}
\label{sec:related_work}

\noindent \textbf{Benchmarking LLMs for socio-scientific domains}
In recent years we have witnessed the emergence of numerous domain-specific benchmarks designed to rigorously evaluate \gls*{llm}s across specialized fields. In the context of environment and environmental sciences, notable contributions include benchmark suite for water engineering~\cite{Yang2024UnlockingTPA}, the ELLE dataset~\cite{Guo2025EnvironmentalLLB}, the EnviroExam benchmark~\cite{Huang2024EnviroExamBEC}, and the WeQA benchmark~\cite{Meyur2024WeQAABD} focusing on \gls*{rag}-based evaluation in the wind energy domain. Furthermore, specialized benchmarks for professional domains have also surfaced such as the LawBench benchmark~\cite{Chen2023LawBenchBLE}, CFinBench~\cite{Guo2024CFinBenchACF}, and TransportationGames~\cite{Zhang2024TransportationGamesBTG}  to evaluate knowledge of \gls*{llm}s in legal, financial and transport domains respectively. SciEval~\cite{Han2023SciEvalAMH} is another example which provides a multi-level evaluation benchmark for scientific research capabilities of LLMs. SciBench~\cite{wang2023scibench} and FrontierMath~\cite{glazer2024frontiermath} are premiere benchmarks to assess scientific and mathematical problem‐solving capabilities of \gls*{llm}s. More recently, PhysReason~\cite{zhang2025physreason}, QASA~\cite{lee2023qasa} and the CURIE dataset~\cite{cui2025curie} have become popular benchmark dataset to evaluate \gls*{llm} performance on scientific reasoning tasks.

% The scientific and mathematical capabilities of large language models have become a focal point of evaluation, with researchers designing increasingly sophisticated benchmarks to evaluate \gls*{llm} reasoning across disciplines.
% \citet{rein2024gpqa} introduced GPQA, a graduate‐level, Google‐resistant question‐answering benchmark designed to test deep comprehension and reasoning. 
% proposed PhysReason, a comprehensive benchmark for physics‐based reasoning. In parallel, \citet{lee2023qasa} introduced QASA for question answering on full scientific articles, and \citet{cui2025curie} released CURIE to evaluate long‐context understanding and multitask scientific reasoning of LLMs.

\noindent \textbf{Frameworks for LLM Evaluations}
% \sameera{It is not clear which corner that you want to emphasize in this section, is it that we are saying the need of the Domain-Specific Evaluation Frameworks?}
% \anurag{No, we're saying there is a need for easy, various context-in-one evaluation framework.}
The evaluation of \gls{llm}s has recently evolved beyond simple task-based metrics toward more sophisticated frameworks that address specific evaluation challenges across diverse domains. More recently, complementary approaches to tackle different aspects of LLM evaluation have emerged~\cite{Yao2024KIEvalAKA,Joty2024ReIFERIC}.
% such as the KIEval~\cite{Yao2024KIEvalAKA} providing a knowledge‐grounded interactive evaluation framework, while \citet{Joty2024ReIFERIC} revisit instruction‐based evaluation using a novel ReIFE methodology. 
% Domain-specific frameworks have also gained prominence, such as the human evaluation framework by~\citet{Sivarajkumar2024AFFB} for healthcare applications derived from systematic literature review.
Efforts to standardize evaluation processes have produced several notable toolkits, including Evalverse~\cite{Kim2024EvalverseUAD}%offering a unified and accessible library for \gls{llm} evaluation
, FMEval~\cite{Franceschi2024EvaluatingLLE}%for systematically benchmarking \gls*{llm} capabilities
, and LLMeBench~\cite{Dalvi2023LLMeBenchAFF}.% providing a flexible framework to accelerate benchmarking. 
For specialized \gls*{rag}-based frameworks, \citet{Guinet2024AutomatedEOG} presents an automated evaluation protocol utilizing task‐specific exam generation. One of the most famous evaluation frameworks is LM Eval Harness~\cite{eval-harness}, which provides a unified platform for measuring LLM performance across hundreds of standardized tasks and has become a cornerstone for reproducible model evaluation in the research community. Similarly, RAGAS~\cite{ragas} provides multiple types of metrics to evaluate LLM responses on dimensions like faithfulness and context relevance, but does not provide a comprehensive workflow framework for end-to-end evaluation. DeepEval~\cite{deepeval} offers testing capabilities for LLM applications but is primarily optimized for OpenAI models, potentially limiting its applicability across different model providers. Our proposed MAPLE framework offers a transparent and modular approach that seamlessly integrates multiple LLM providers, supports customizable prompt templates, handles various context types, and incorporates diverse evaluation metrics within a unified evaluation pipeline.

\section{Discussion and Conclusion}
\label{sec:conclusion}
In this study, we conduct the initial investigation into the performance of \gls*{llm}s within the \gls*{nepa} domain and its associated documents. To facilitate this, we introduce NEPAQuAD, a question-and-answering benchmark designed to evaluate a model's capability to understand the legal, technical, and compliance-related content found in \gls*{nepa} documents. We assess five long-context \gls*{llm}s (both closed- and open-source) designed for handling extensive contexts across various contextual settings. 

Our analysis indicates that \gls*{nepa} regulatory reasoning tasks pose a significant challenge for \gls*{llm}s, particularly in terms of understanding the complex semantics and effectively processing the lengthy documents. The findings reveal that models augmented with the \gls*{rag} technique surpass those that are simply provided with the PDF content as long context. This suggests that incorporating more relevant knowledge retrieval processes can significantly enhance the performance of \gls*{llm}s on complex document comprehension tasks like those found in the \gls*{nepa} domain. In addition, we noticed that these models struggle to use long input contexts to answer more difficult questions that require multiple steps of reasoning.  For example, models performed best in answering closed questions and worst in answering divergent and problem-solving questions. However, when these models are augmented with additional context, as seen in the \gls*{rag} setup, their performance can be significantly improved. %We encourage the research community to focus on advancing the complementary use of long-context \gls*{llm}s and \gls*{rag} techniques together, rather than treating them as distinct, separate approaches.
\section{Limitations}% and Future Work}
\label{sec:limitations}

As with all work, our proposed system for EIS long documents also has some limitations. We list those limitations as follows:

% \textbf{Lack of Fine-tuning Process.} In this work, we use the original embedding model BGE \cite{bge_embedding} for silver passages suggestion. Given the fact that optimization on fine-tuning is a very important part of strategies of improving performance of IR models, as shown in \cite{horawalavithana2023scitune}, we will evaluate our work with fine-tuned models for our domain of EIS documents' questions and answers in future work.

\noindent\textbf{Restriction of token limitation on full PDF context.} 
While we are able to use the Gemini model with token length as 1.5 million, we could only use 128K tokens per query for response generation with Claude and GPT-4. Thus, we needed to truncate the content of Full PDF to run these two models. This might cause the performance drop on the full PDF context mode. In future work, we should analyze more carefully about the impact of token truncation.

\noindent\textbf{Uncertainty of generated responses by \gls{llm}s.} Due to budget constraints, we conducted only one phase of response generation across different configurations. This introduces a risk of uncertain outputs, as LLMs might generate different responses each time, even with the same input, as demonstrated in another study \cite{wagle2024empirical}. Multiple runs of the same model might mitigate this effect.

\noindent\textbf{Challenges of human judgment.}
Currently, we leverage human evaluation only on a subset of entries in the benchmark. With sufficient time and resources, a thorough evaluation of the entire dataset following the same methodology will likely make the dataset even more robust.
% In future work, we plan to involve more NEPA experts in a more systematic manner to expand the dataset with human judgment results and to perform proper adjudication meetings between NEPA experts to reconcile conflicting results.

\noindent\textbf{Imbalance of question types.}
In the current benchmark, just around half the questions are closed type. This is due to a myriad of reasons, including difficulty in generating open questions and the time-consuming nature of open question evaluation. While open-questions do take significantly more human review and effort, a future benchmark with even more such questions will undoubtedly provide an even tougher challenge to LLMs.

\noindent\textbf{Random errors when using RAGAS}
We noticed that RAGAS evaluation framework~\cite{es2023ragas} produces random errors and "NaN" values in specific types of question. While running experiments to analyze the details of this fall outside of scope of this work, our preliminary tests lead us to assume that it might be due to the way the framework interacts with generative models and tries to conform their output to its formats.

\noindent\textbf{Bias in automated evaluation}
There might be a potential bias in the answer correctness evaluation process due to the use of GPT-4 to assess the outputs of various models.
There is a concern that GPT-4 may inherently prefer the outputs generated by the same model over others in the factual correctness evaluation. 
This could lead to skewed evaluation results, where GPT-4's outputs are rated more favorably, not necessarily because they are superior, but because of the inherent biases in the evaluation model (GPT-4).

To address the potential bias in the answer correctness evaluation process, we assess both factual and semantic correctness in the evaluation. For semantic correctness, we utilize the BGE~\cite{bge_embedding} as the embedding model and we calculate the semantic similarity between the model's outputs and the ground-truth answers independently of GPT-4's own evaluation mechanisms. By combining both factual and semantic correctness, we aim to accurately reflect the true performance of various models, including GPT-4.

\section*{Acknowledgement}
This work was supported by the Office of Policy, U.S. Department of Energy, and Pacific Northwest National Laboratory, which is operated by Battelle Memorial Institute for the U.S. Department of Energy under Contract DE-AC05–76RLO1830.  This paper has been cleared by PNNL for public release as PNNL-SA-211465.

\bibliographystyle{ACM-Reference-Format}
\bibliography{custom}

\appendix
% \newpage

\section{Prompt Template}
\label{app:prompt}

The template for the prompt is shown below:

\prompt{1. Generate \{placeholder for number of questions\} \{placeholder for type of question\} questions from the following context.\\
Context: \{placeholder for gold passage\}\\
2. Definition of \{placeholder for type of question\} question: \{placeholder for definition provided by NEPA experts\}\\
3. The questions can be similar to following sample questions:\\
{placeholder for sample questions provided by NEPA experts}\\
4. Return the questions in CSV format with columns question / answer / proof.}
We provided appropriate placeholders for the type and number of questions to be generated, the excerpt from which the question will be generated along with the sample questions of the same type as provided by the \gls*{nepa} experts.
We restricted the output for each prompt in a CSV format with three fields: question, answer, and proof. The `proof' column stored the part of the gold passage that the model picked as evidence for the provided answer to the question.

\section{Question Definitions}
\label{sec:question_defs}
NEPA experts reviewed and created the definitions for each question types as following.
\begin{enumerate}
    \item \textbf{Closed questions:} Closed questions have two possible answers depending on how you phrase it: “yes” or “no” or “true” or “false.” 
    \item \textbf{Comparison questions:} Comparison questions are higher-order questions that ask listeners to compare two things, such as objects, people, ideas, stories or theories.
    \item \textbf{Convergent questions:} convergent questions are designed to try and help you find the solution to a problem, or a single response to a question.
    \item \textbf{Divergent questions:} Divergent questions have no right or wrong answers but rather encourage open discussion. While they are similar to open questions, divergent questions differ in that they invite the listener to share an opinion, especially one that relates to future possibilities.
     \item \textbf{Evaluation questions:} Evaluation questions, sometimes referred to as key evaluation questions or KEQs, are high-level questions that are used to guide an evaluation. Good evaluation questions will get to the heart of what it is you want to know about your program, policy or service.
     \item \textbf{Funnel questions:} Funnel questions are always a series of questions. Their sequence mimics a funnel structure in that they start broadly with open questions, then segue to closed questions.
     \item \textbf{Inference questions:} Inference questions require learners to use inductive or deductive reasoning to eliminate responses or critically assess a statement.
     \item \textbf{Problem-solving questions.} Problem-solving questions present students with a scenario or problem and require them to develop a solution. 
     \item \textbf{Process questions:} A process question allows the speaker to evaluate the listener's knowledge in more detail.
     \item \textbf{Recall questions:} A recall question asks the listener to recall a specific fact.
\end{enumerate}

\section{Sample Questions}
\label{sec:question_samples}
In this sections, we listed the sets of sample questions we used for each types of questions.
\subsection{Closed questions}
\begin{itemize}
    \item Are there any federally recognized Tribes in a 50-mile radius of [PROJECT]?
 \item Are there any federally recognized species of concern in a 50-mile radius of [PROJECT]?
 \item Did [AGENCY] approve the licensing action
 \item Did the EIS consider [SUBJECT]?
\end{itemize}
\subsection{Comparison questions}
\begin{itemize}
    \item Which Tribes were consulted in [PROJECT 1] and not [PROJECT 2] and vice-versa?
 \item What are some differences between [STUDY 1] and [STUDY 2] that might account for differences in species count for [SPECIES]?
 \item Compare the considered alternatives in [PROJECT 1] with those in [PROJECT 2].
 \item Compare the outcomes of surveys from the new reactor EIS with the license renewal EIS for [RPOJECT].

\end{itemize}
\subsection{Convergent questions}
\begin{itemize}
    \item Which other species of concern could logically be in within the 50-mile radius around the [PROJECT]?
 \item How many similar projects could be built before the impact level for air quality was rated as high?
 \item If the area of effect for the proposed action were increased by 50\%, what additional federal species of concern would need to be addressed?

\end{itemize}
\subsection{Divergent questions}
\begin{itemize}
    \item What considerations should the [AGENCY] addressed in the document but didn't?
\end{itemize}
\subsection{Evaluation questions}
\begin{itemize}
    \item Based on NEPA evaluations done in the vicinity of [PROJECT], does the conclusion of the Historical and Cultural resources section appropriately weigh the concerns of Tribal leaders?
 \item Extrapolating using this and other NEPA evaluations, what is the long term outlook for [SPECIES] in the vicinity?
 \item How have [AGENCY'S] NEPA reviews trended over time and would this review have the same outcome 10 years ago or 10 years from now?
 \item In the license renewal EIS for [PROJECT], which impacts have changed from the initial EIS and why?

\end{itemize}
\subsection{Funnel questions}
\begin{itemize}
    \item Which federally recognized Tribes are in a 50-mile radius of [PROJECT]? Which Tribes participated in this EIS? What were the concerns fo participating Tribes? What mitigations were made? 
 \item Which federally recognized species of concern are in a 50-mile radius of [PROJECT]? What mitigations, if any, were made to project those species?
 \item Which alternatives were discussed? Which were considered? Why was [ALTERNATIVE] not considered?
 \item Which resource areas were discussed in the Affected Environmnent section of the document? 
 \item What were the impacts of the proposed action on [SUBJECT]? 
 \item Did [AGENCY] consider [X] when evaluting [SUBJECT]?

\end{itemize}
\subsection{Inference questions}
\begin{itemize}
    \item If the federally recognized [TRIBE] has land in the vicinity of [PROJECT 1] like it does in the vicity of [PROJECT 2], what concerns might [TRIBE] have with [PROJECT 1]?
    \item If the primary migitation for [SPECIES] for [PROJECT TYPE] in the past has been [MITIGATION], what would you expect the mitigation to be for [PROJECT]?
 \item If [AGENCY 1] and [AGENCY 2] typically agree on impact levels and [AGENCY 1] found large impact in terrestrial ecology for an action in a nearby area, what would [AGENCY 2] find?
 \item If mitigations for air quality for [PROJECT 1] were effective and the same mitigations were applied to [PROJECT 2], what would we assume the outcome to be for [PROJECT 2]?
\end{itemize}
\subsection{Problem-solving questions}
\begin{itemize}
    \item Given the following references, evaluate the effect of a new nuclear plant at [SITE] on cultural and historic resources in the vicinity.
 \item Given the location of the [PROJECT], create a list of aquatic species likely present in a 50-mile radius.
 \item Write an Abstract for [PROJECT]
 \item Given the list of reference in [SECTION] of [PROJECT 1] create a list of references applicable to [PROJECT 2]. Provide hyperlinks and ML numbers, if available.
\end{itemize}
\subsection{Process questions}
\begin{itemize}
    \item How does this document define the NEPA process for consultation with Tribes?
 \item How does [AGENCY] define the area of effect for the proposed action?
\end{itemize}
\subsection{Recall questions}
\begin{itemize}
    \item What references did [AGENCY] use in evaluating the effect of the applicant's proposed action on [SPECIES]?
    \item Which resource areas indicated a moderate or large impact due to the proposed action?

\end{itemize}

\section{Human Annotation Process}
\label{sec:human-annotation-example}

The first example shows a triplet which was accepted by both \gls*{nepa} experts (shown in green), whereas the second example shows an example where there was a conflict of opinions between the \gls*{nepa} experts (shown in orange).

\goodeval{
\textbf{Type}: process\\
\textbf{Question}: How does the document define the socioeconomic conditions for the analysis of the human environment?\\
\textbf{Answer}: The document defines socioeconomic conditions for the analysis as indicators including population, employment, unemployment rate, income, cost of living, and housing availability.\\
\textbf{Proof}: Socioeconomics pertains to the social and economic conditions of the human environment. For this analysis, the indicators of socioeconomic conditions include population, employment, unemployment rate, income, cost of living, and housing availability.\\
\textbf{\gls*{nepa} expert 1 evaluation}: Correct triplet.\\
\textbf{\gls*{nepa} expert 2 evaluation}: The triplet is correct.\\
\textbf{Overall decision}: Include in \nepaquad %since both experts accepted the triplet.
}

\eval{
\textbf{Type}: Divergent\\
\textbf{Question}: What future cultural or commercial impacts might arise from the heat and electrical generation and distribution facility upgrades at Fort Wainwright, considering its status as a regional hub?\\
\textbf{Answer}: Potential impacts could range from attracting new businesses and boosting the local economy to affecting the cultural demographics and social dynamics within Fort Wainwright and the surrounding villages, given its role as the cultural and commercial center of the Interior Region.\\
\textbf{Proof}: The borough is the cultural and commercial center of the Interior Region as well as a hub for villages located hundreds of miles outside the region.\\
\textbf{\gls*{nepa} expert 1 evaluation}: The type of generated question is correct. The answer and proof are incorrect. Table ES-1 in the document provides a summary of socioeconomic impacts that would improve the answer. The answer also conflates Fort Wainwright as the regional hub, whereas the actual text of the EIS identifies the FNSB as the cultural hub.\\
\textbf{\gls*{nepa} expert 2 evaluation}: Provided triplet is correct.\\
\textbf{Overall decision}: Consult with a third \gls*{nepa} expert. %to evaluate the quality of the triplet.
}

\section{EIS Dataset}
Table~\ref{tbl:dataset:statistics} reports the statistics of the EIS data that used to create the benchmark.

\begin{table*}[!t]
\caption{Statistics on the EIS documents used in the evaluation}
\begin{center}

    % \setlength\tabcolsep{1pt}
    % \resizebox{\columnwidth}{!}{
    \small
\begin{tabular}{|p{0.4\linewidth}|p{0.3\linewidth}|p{0.1\linewidth}|p{0.1\linewidth}|}
\hline
\multicolumn{1}{|c|}{\textbf{Document Title}}             & \textbf{Agency} &  \textbf{\#Pages} & \textbf{\#Tokens} \\ \hline
Continental United States Interceptor Site        & Missile Defense Agency, Department of Defense                              & 74                                          & 41,742                                       \\ \hline
Supplement Analysis of the Final Tank Closure
and Waste Management for the Hanford Site, Richland,
Washington, Offsite Secondary Waste Treatment and
Disposal & Hanford Site Office, Department of Energy                    & 63                                          & 43,167                                       \\ \hline
Nationwide Public Safety Broadband Network
Final Programmatic Environmental Impact Statement
for the Southern United States                  &      Department of Commerce                                            & 86                                          & 43,985                                       \\ \hline
T-7A Recapitalization at Columbus Air Force Base, Mississippi           &  United States Department of the Air Force (DAF), Air Education and
Training Command (AETC).                                       & 472                                         & 179,697                                      \\ \hline
Oil and Gas Decommissioning Activities on the Pacific Outer Continental Shelf             &         The Bureau of Safety and Environmental Enforcement (BSEE) and Bureau of Ocean
Energy Management (BOEM)                           & 404                                         & 271,545                                      \\ \hline
Final Environmental Impact Statement
for the Land Management Plan
Tonto National Forest          &           Department of Agriculture, Forest Service                                & 472                                         & 325,641                                      \\ \hline
Final Environmental Impact Statement for Nevada Gold Mines LLC's Goldrush Mine Project, Lander and Eureka Counties, NV             &     Bureau of Land Management, Interior.            & 454                                         & 413,083                                      \\ \hline
Addressing Heat and Electrical Upgrades at Fort Wainwright, Alaska             &       Department of the Army, Department of Defense                                & 618                                         & 514,003                                      \\ \hline
Sea Port Oil Terminal Deepwater Port Project        &       The U.S. Coast Guard (USCG) and Maritime Administration (MARAD), Department oF Transportation                                      & 890                                         & 613,214                                      \\ \hline

\end{tabular}
% }
    
\end{center}
\label{tbl:dataset:statistics}
\end{table*}

\section{Examples from NEPAQuAD}
\label{sec:app-gold}
Here we present an example of a gold passage which was selected by \gls{nepa} experts from an \gls{eis} document on the Fort Wainwright generation and distribution facility.

\context{3.5.1.1 Definition of Resource
Socioeconomics pertains to the social and economic conditions of the human environment. For this analysis, the indicators of socioeconomic conditions include population, employment, unemployment rate, income, cost of living, and housing availability. Current data on these indicators for the affected environment would provide the baseline information on the socioeconomic well-being of the local areas/region upon which potential effects of the proposed project alternatives are compared. The proposed heat and electrical generation and distribution facility upgrades would occur within Fort Wainwright, located in the FNSB. The borough is the cultural and commercial center of the Interior Region as well as a hub for villages located hundreds of miles outside the region (Alaska Department of Labor and Workforce Development [ADOLWD] 2018).
}

Now we present some examples of questions from NEPAQuAD which have been generated using our methodology in Section~\ref{sec:benchmark} and accepted to be included by the \gls*{nepa} experts. These examples include the question-answer-proof triplets which are generated using our methodology.
\begin{itemize}
    \item \textbf{Closed}~question: `Does the definition of resource include biological studies?', answer: `No', proof: `Socioeconomics pertains to the social and economic conditions of the human environment.'
    \item \textbf{Comparison}~question: `How do the roles of Fort Wainwright and the FNSB differ with respect to their regions in Alaska?', answer: `Fort Wainwright is a military facility undergoing upgrades within the FNSB, which is the cultural and commercial center of the Interior Region and acts as a hub for remote villages. Fort Wainwright is located within the FNSB but caters specifically to military functions, while the FNSB serves a wider role in the region.', proof: `The proposed heat and electrical generation and distribution facility upgrades would occur within Fort Wainwright, located in the FNSB. The borough is the cultural and commercial center of the Interior Region as well as a hub for villages located hundreds of miles outside the region.'
    \item \textbf{Divergent}~question: `What future cultural or commercial impacts might arise from the heat and electrical generation and distribution facility upgrades at Fort Wainwright, considering its status as a regional hub?', answer: `Potential impacts could range from attracting new businesses and boosting the local economy to affecting the cultural demographics and social dynamics within Fort Wainwright and the surrounding villages, given its role as the cultural and commercial center of the Interior Region.', proof: `The borough is the cultural and commercial center of the Interior Region as well as a hub for villages located hundreds of miles outside the region.'
    \item \textbf{Funnel}~question: `What is the role of the FNSB as mentioned in the document?', answer: `The FNSB is the cultural and commercial center of the Interior Region and a hub for villages located hundreds of miles outside the region.', proof: `The borough is the cultural and commercial center of the Interior Region as well as a hub for villages located hundreds of miles outside the region.'
    \item \textbf{Inference}~question: `If Fort Wainwright's proposed project serves as a cultural and commercial hub for outlying villages, how might the facility upgrades affect these external areas?', answer: `The facility upgrades could potentially affect the socioeconomic well-being of the villages that relay on Fort Wainwright as a hub, due to interconnected economic and cultural activities.', proof: `The borough is the cultural and commercial center of the Interior Region as well as a hub for villages located hundreds of miles outside the region.'
    \item \textbf{Process}~question: `What process is outlined for establishing the baseline information on the socioeconomic well-being of the local areas/region affected by the proposed project?', answer: `The process for establishing the baseline information involves collecting current data on indicators such as population, employment, unemployment rate, income, cost of living, and housing availability for the affected environment. This baseline is used for comparing the potential effects of the proposed project alternatives.', proof: `Current data on these indicators for the affected environment would provide the baseline information on the socioeconomic well-being of the local areas/region upon which potential effects of the proposed project alternatives are compared.'
\end{itemize}

\section{\maple Modules}
\label{ref:app-sec-maple}
\subsection{LLM Handlers}
The LLM Handler infrastructure provides a unified interface for interacting with various Language Model providers. 
This module implements an abstract base class that standardizes the interaction with different LLM services while accommodating provider-specific requirements.
The key features of the LLM Handler module are discussed below.

\paragraph{Supported providers.}
The LLM handler of MAPLE supports the major cloud providers:
\begin{itemize}
    \item Azure OpenAI services with GPT models
    \item AWS Bedrock with Claude, LLama and Mistral models
    \item Google Vertex AI with Gemini models
    \item Local deployment of HuggingFace models
\end{itemize}
    
\paragraph{Authentication management}
The authentication management system in MAPLE's LLM Handlers provides provider-specific authentication while maintaining consistent security standards. 
Azure OpenAI services utilize API key validation and endpoint configuration, while AWS Bedrock supports both role-based credentials through AWS STS and profile-based authentication through AWS SSO. 
Google Vertex AI implements service account authentication using JSON key files with automatic credential refresh mechanisms. 
For local HuggingFace models, there is no authentication system, rather it only requires the model path.
The system implements credential caching, automatic retries for authentication failures, and secure credential storage across all handlers, effectively abstracting provider-specific authentication complexities from the evaluation pipeline.
    
\paragraph{Response generation capabilities}
The LLM Handlers provide a standardized interface for generating responses across different providers while maintaining provider-specific optimizations.
All handlers implement consistent response formatting, error handling for failed generations, and proper resource cleanup, ensuring reliable response generation regardless of the underlying provider.

\subsection{Utility Modules}
The Utility Modules provide essential support functions for data management, logging, and visualization. These modules form the foundation for data processing and result analysis within the framework.

\paragraph{Data Loading and Processing}
The data loading system provides structured handling of benchmark datasets and evaluation results. 
A key feature is its automatic context type detection based on CSV column availability: presence of the `file\_name' column enables both RAG and PDF context types as it allows document linkage, while the `context' column enables Gold context type by providing ground truth contexts. 
The basic 'none' context type is always supported as it requires only question and answer columns. 
The system enforces data validation for required fields and handles optional attributes flexibly. 
It processes both input benchmark questions and evaluation results, maintaining data consistency and providing clear feedback on data quality issues through a hierarchical exception handling system. 
This context type detection allows downstream components to automatically determine valid evaluation strategies based on the available data.

\paragraph{Logging Infrastructure}
The logging infrastructure implements a centralized logging system through a singleton pattern, ensuring consistent log management across all components. 
It provides configurable log levels, automatic file rotation, and structured log formatting. 
The system maintains separate logs for different components while enabling unified log access, with automatic creation of log directories and files, and proper handling of concurrent logging requests.

% \paragraph{Visualization Tools}
% The visualization system offers two primary tools: a heatmap generator for comparative analysis and a bar plot generator for metric visualization. 
% The heatmap generator supports multi-context performance visualization and question type comparisons, while the bar plot generator handles metric distribution analysis and model performance comparisons. 
% Both tools implement customizable styling options and automatic figure management.

\paragraph{Configuration Management}
The configuration system handles various framework settings through a combination of YAML files and environment variables. 
It manages provider credentials, folder paths, and evaluation parameters, implementing validation checks and providing default values where appropriate. 
The system supports different configurations for various evaluation scenarios while maintaining security for sensitive information.

\paragraph{Type Validation}
The type validation system enforces data consistency through strongly-typed data structures. It implements comprehensive validation for both required and optional fields, handles missing values appropriately, and provides clear error messages for validation failures. The system ensures data integrity across the framework while maintaining flexibility for different data formats.

\subsection{Evaluator Module}
The Evaluator Module manages the response generation process, implementing different context types and handling the evaluation pipeline. This module coordinates the interaction between LLM handlers and input data.

\paragraph{Context Type Management}
The Evaluator module supports multiple context types for response generation, each with specialized processing pipelines. 
The supported context types are discussed here.
\begin{itemize}
    \item \texttt{none} context type: The basic mode generates responses without additional context. 
    \item \texttt{pdf} context type: The `pdf' context mode processes entire documents, managing token limits and content extraction. 
    The text chunks are loaded from the JSON files stored within the data directory (required input for this context type).
    In case of tokens exceeding the token limit, we define methods to truncate the text chunks to stay within the allowed limits.
    \item \texttt{rag} context type: This mode integrates with a vector database for relevant document retrieval and filtering. 
    The path to the vector database and the collection name needs to be provided as inputs for this context type.
    \item \texttt{gold} context type: This mode utilizes predefined ground truth contexts for controlled evaluation scenarios. 
\end{itemize}
The system automatically validates context availability and requirements for each mode, ensuring appropriate context handling based on available data.

\paragraph{Prompt Template Management}
The Evaluator module supports customizable prompt engineering through a template-based system. 
Users can provide prompt templates with placeholders for 'question' and 'context' (when applicable) through a template file and provide its path as an input. 
The system automatically formats the prompt by replacing these placeholders with actual content during evaluation, raising appropriate warnings if required placeholders are missing for the chosen context type. 
For instance, templates for 'gold' context type must contain both {question} and {context} placeholders, while basic evaluation might only require {question}. 
This flexibility in prompt design allows users to experiment with different prompting strategies and maintain consistent prompt structures across evaluations without modifying the core pipeline. 

\paragraph{Progress Management and Persistence}
The evaluator implements a robust progress tracking and persistence system that processes benchmark questions sequentially while maintaining state. 
For each benchmark question, it generates a response, processes the result, and immediately stores it in the output CSV file. 
This immediate persistence strategy ensures that progress is never lost due to interruptions. 
When restarting an interrupted evaluation, the system automatically detects previously processed questions by checking the existing output file and continues from the last successful evaluation. 
This approach provides resilience against system failures and allows for long-running evaluations to be safely paused and resumed, while maintaining detailed logs of each processing step.

\paragraph{Response Management}
The response management system handles the collection, validation, and storage of generated responses with their associated metadata. 
For RAG-based evaluations, it tracks source documents and relevance scores, while for context-based generations, it maintains context-response relationships. 
The system implements structured storage of responses, supporting both incremental updates and batch commits, with comprehensive logging of generation parameters and outcomes.

\paragraph{Error Recovery}
The error recovery system provides robust handling of various failure scenarios during evaluation.
It implements automatic retries for transient failures, and structured logging of unrecoverable errors. 
The system maintains evaluation progress despite individual question failures and provides detailed error reporting for post-evaluation analysis.

\subsection{Metrics Module}
The Metrics Module implements RAGAS evaluation metrics for assessing the quality of LLM responses. This module provides comprehensive evaluation capabilities across multiple dimensions of response quality.

\paragraph{Supported Metrics}
The Metrics module implements a subset of RAGAS evaluation metrics for assessing LLM response quality. 
The current version supports five core metrics: answer correctness for measuring accuracy, answer similarity for response alignment, context precision and recall for evaluating context relevance, and answer faithfulness for assessing response consistency with provided contexts. 
Each metric provides a score between 0 and 1, enabling quantitative assessment of different response aspects.

\paragraph{Batch Evaluation}
The module processes multiple evaluation responses in configurable batch sizes to optimize resource usage and evaluation speed. 
For each batch, it computes all selected metrics simultaneously using the RAGAS framework, managing memory efficiently by processing fixed-size subsets of the full evaluation dataset. 
This batched approach provides a balance between processing speed and resource consumption.

\paragraph{Progress Management}
Similar to the Evaluator module, the Metrics module implements immediate result persistence by writing metric scores to the output CSV file after each batch evaluation. 
The system tracks progress through the evaluation set and supports continuation from the last successful evaluation in case of interruptions. 
This ensures reliable processing of large evaluation datasets while maintaining evaluation state.

\paragraph{NaN Recovery and Recomputation}
The module provides functionality to identify and recompute metrics for responses where RAGAS evaluation resulted in NaN (Not a Number) values. 
It scans the results CSV file for NaN entries in specified metric columns, extracts the corresponding responses and contexts, and performs targeted re-evaluation of these specific entries. 
The recomputed values are then seamlessly integrated back into the original results file, maintaining data consistency while improving evaluation completeness.

\paragraph{Result Storage and Visualization}
Evaluation results are stored in structured CSV files with consistent column ordering: file metadata columns followed by question details, expected and predicted responses, and individual metric scores. 
This standardized format enables direct integration with the visualization utilities, supporting generation of comparative heatmaps and performance bar plots.
The results can be analyzed across different models, context types, and question categories using the visualization tools available within the utilities module of MAPLE.

\end{document}